\documentclass[wcp]{jmlr}
\jmlrvolume{1}
\jmlryear{2017}
\jmlrworkshop{ICML 2017 AutoML Workshop}

\usepackage[utf8]{inputenc} 
\usepackage[T1]{fontenc}    
\usepackage{hyperref}       
\usepackage{url}            
\usepackage{booktabs}       
\usepackage{amsfonts}       
\usepackage{nicefrac}       
\usepackage{microtype}      

\usepackage{amsmath}
\usepackage{xcolor}
\usepackage{graphicx}
\usepackage{caption}

\newcommand{\norm}[1]{\left\lVert#1\right\rVert}
\newcommand{\Perp}[0]{\ensuremath{Perp} }

\newcommand{\comment}[1]{}

\title[Auto-Perplexity t-SNE]{Automatic Selection of t-SNE Perplexity}

  \author{\Name{Yanshuai Cao}${}$  \Email{yanshuai.cao@rbc.com}\and
    \Name{Luyu Wang}${}$ \Email{luyu.wang@rbc.com}\\
    \begin{center}
          \vspace{-.6cm}
  \addr RBC Research Institute
  \end{center}
    }

\begin{document}

\maketitle
\vspace{-1.5cm}
\begin{abstract}
  t-Distributed Stochastic Neighbor Embedding (t-SNE) is one of the most widely used dimensionality reduction methods for data visualization, but it has a perplexity hyperparameter that requires manual selection. In practice, proper tuning of t-SNE perplexity requires users to understand the inner working of the method as well as to have hands-on experience. We propose a model selection objective for t-SNE perplexity that requires negligible extra computation beyond that of the t-SNE itself. We empirically validate that the perplexity settings found by our approach are consistent with preferences elicited from human experts across a number of datasets. The similarities of our approach to Bayesian information criteria (BIC) and minimum description length (MDL) are also analyzed.
\end{abstract}
\begin{keywords}
t-SNE, perplexity, hyperparameter tuning, Bayesian information criteria
\end{keywords}

\section{Introduction}
\vspace{-.1cm}
t-Distributed Stochastic Neighbor Embedding (t-SNE) \citep{maaten2008visualizing,van2014accelerating} is arguably the most widely used nonlinear dimensionality reduction method for data visualization in machine learning and data science.
Using t-SNE requires tuning some hyperparameters, notably the perplexity. Although according to \citet{maaten2008visualizing}, t-SNE results are robust to the settings of perplexity, in practice, users would still have to interactively choose perplexity by visually comparing results under multiple settings. Often, complete novice users potentially need practice tuning t-SNE on various simple problems in order to gain enough insight and skills to use it properly \citep{wattenberg2016use}. The lack of automation in selecting this crucial hyperparameter poses difficulty for non expert users who do not understand the inner working of the t-SNE algorithm, and could lead to misinterpretation of data. In this work, we propose an approach to automatically set perplexity, which requires no significant extra computation beyond runs of t-SNE optimization. The proposed approach is based on an objective that is function of perplexity and resulting KL divergence of learned t-SNE. We motivate the novel objective from the perspective of model selection and validate it by showing that its minimum agrees with human expert selection in empirical studies.

\section{t-Distributed Stochastic Neighbor Embedding}
\vspace{-.1cm}
t-SNE tries to preserve local neighborhood structure from high dimensional space in low dimensional space by converting pairwise distances to pairwise joint distributions, and optimize low dimensional embeddings to match the high and low dimensional joint distributions. Specifically, let $\{x_i\}^n_{i=1}$ be high dimsensional data points, and $\{y_i\}^n_{i=1}$ the corresponding low dimensional embedding points, t-SNE defines joint distribution of point $i,j$ as follows:
the low dimensional joint distribution is
\begin{equation}
  q_{ij} = {\left(1+\norm{y_i - y_j}^2\right)^{-1}} \Bigg/ {\sum_{s \neq t} \left(1+\norm{y_s - y_t}^2\right)^{-1}} 
\end{equation}
and the high dimensional one is defined as symmetrized conditionals: $p_{ij} = {(p_{i|j} + p_{j|i})}/{2n}$, where 
\begin{equation}
p_{i|j} = \frac{\exp(- \norm{x_i - x_j}^2 / 2\sigma_j^2)}{\sum_{s \neq j} \exp(- \norm{x_s - x_j}^2 / 2\sigma_j^2)} \\ \label{cond_p_ij}
\end{equation}
Finally, the t-SNE optimizes $\{y_i\}_i$ to minimize the Kullback–Leibler divergence from low dimensional distribution $Q$ to high dimensional $P$:
\begin{equation}
  \text{KL}(P||Q) = \sum_{i \neq j} p_{ij} \log \frac{p_{ij}}{q_{ij}} \label{kl_obj}
\end{equation}
\vspace{-1cm}
\subsection{Perplexity}
In Eq.\ \ref{cond_p_ij} includes $\sigma_j$ which defines the local scale around $x_j$. The value for $\sigma_j$ is not optimized or specified by hand individually, but rather found by bisection search to match a pre-specified perplexity value \Perp. The perplexity of $p_j$ is $Perp(p_j) = 2 ^ {H(p_j)}$, where $H(P_j) = - \sum_j p_{i|j} \log_2 p_{i|j}$, and $\sigma_j$ is selected so that $Perp(p_j) = \Perp$.

\Perp is a hyperparameter of the t-SNE algorithm and is central to what structure t-SNE finds. Larger \Perp leads to larger $\sigma_j$ across the board, so that for each data point, more neighbours have significant $p_{i|j}$.
\section{Automatic selection of perplexity}
\begin{figure}[h]
  \centering
  \begin{subfigure}{\label{fig:kl}}
      \centering
  \includegraphics[width=2.4in]{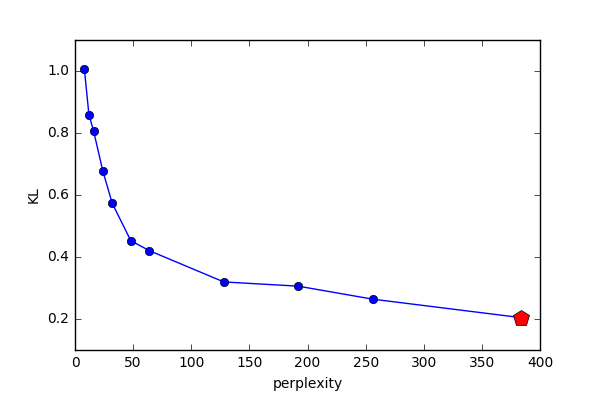} 
  \end{subfigure}  
  \begin{subfigure}{\label{fig:pbic}}
      \centering
  \includegraphics[width=2.4in]{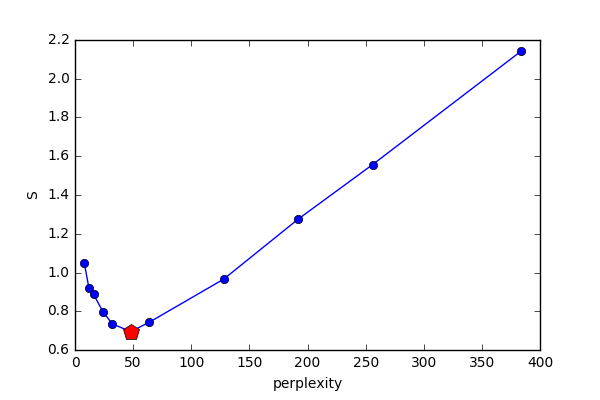}
  \end{subfigure}

  \begin{subfigure}{\label{fig:kl_y}}
      \centering
  \includegraphics[width=2.4in]{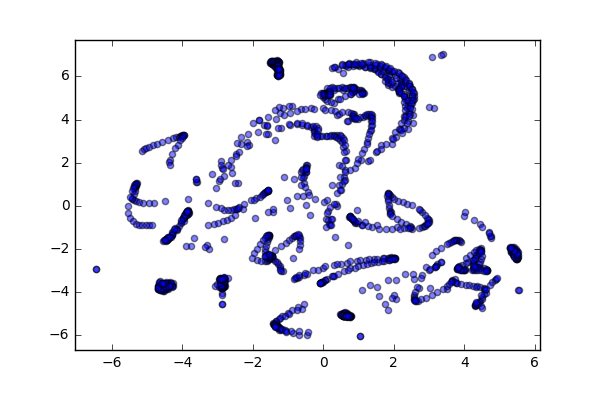}
  \end{subfigure}  
  \begin{subfigure}{\label{fig:pbic_y}}
      \centering
  \includegraphics[width=2.4in]{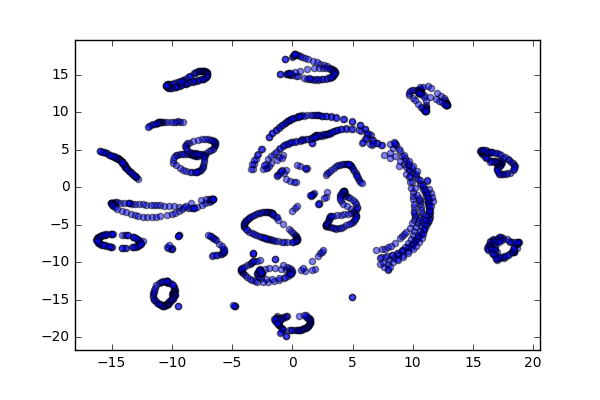}
  \end{subfigure}
  
  \caption{KL divergence (\ref{fig:kl}) and $S$ (\ref{fig:pbic}) as function of \Perp on Coil20 dataset, along with t-SNE maps (\ref{fig:kl_y} and \ref{fig:pbic_y}) at their respective $\text{argmin}$ locations marked by red markers.}  
\end{figure}

The value of KL divergence from different perplexities cannot be compared to assess the quality of embeddings,  since the final KL divergence typically decreases as perplexity increases, as illustrated in Fig.\ \ref{fig:kl}, so that model selection based on KL divergence alone will always lead to very large \Perp. However, the resulting embeddings from large \Perp are usually suboptimal in capturing the underlying pattern of the data, as demonstrated in Fig.\ \ref{fig:kl_y}. In the limit, for \Perp equal to the number of data points, the resulting embeddings usually form a Gaussian or uniform like blob and completely fails to capture any interesting structure.  This suggests that trading off between the final KL divergence and \Perp could potentially lead to good embeddings. Based on this intuition, we design the following criteria:
\vspace{-.1cm}
\begin{equation}
  S(\Perp) =  2\text{KL}(P||Q) + \log(n) \frac{\Perp}{n} \label{pbic}
\end{equation}

Corresponding to $\text{KL}$ in Fig.\ \ref{fig:kl}, $S$ as a function of \Perp is illustrated in Fig.\ \ref{fig:pbic}. To automatically set \Perp, we can perform derivative free optimization of $S$ with respect to \Perp, for instance with Bayesian optimization \citep{brochu2010tutorial} if each t-SNE takes a long time, or  simply grid search if computational cost is low. Implicit in our proposal is that t-SNE has to find the optimal $Q$ given a particular \Perp. In practice, poor convergence of the optimization would affect the final values of $KL$, and hence could potentially impact the result of automatic \Perp tuning. In practice however, we find that the default values of t-SNE optimization in \cite{maaten2008visualizing} allows sufficient consistency in convergence to support robust \Perp selection via Eq.\ \ref{pbic} in a wide range of problems.

In Sec.\ref{experiments} we will demonstrate that $\Perp$ that minimizes $S$ agrees with selection by human users across a number of datasets. But before that, we will motivate Eq.\ \ref{pbic} by relating it to Bayesian Information Criteria (BIC), and minimizing description length.

\subsection{Interpretation as reverse complexity tuning via pseudo BIC (pBIC)}

Eq.\ \ref{pbic} bears resemblance to Bayesian Information Criteria (BIC) \citep{schwarz1978estimating}:
\begin{equation}
  \text{BIC} = -2\log(\hat{L}) + \log(n)k
\end{equation}
where the first term $-2\log(\hat{L})$ is goodness-of-fit of the maximum-likelihood-estimated model ($\hat{L}$), while the second term $\log(n)k$ controls the complexity of the model by penalizing the number of free parameters $k$ scaled by $\log(n)$. Although we do not have a formal derivation of Eq.\ \ref{pbic} similar to BIC derivation as a large sample approximation to the negative marginal log likelihood, there is strong parallel between the two, in terms of both their forms and behaviours of balancing data-fit and complexity. 

The terms in Eq.\ \ref{pbic} are analogous to those of BIC, but the way the complexity changes is reversed: instead of increasing complexity of model to fit data better, increasing \Perp reduces complexity of the pattern in data to be modelled, so that the same lower dimensional space can embed them better. This is because when projecting from high to low dimensional spaces, there is not enough ``room'' in lower dimensional space to preserve all structure in high dimension, i.e. the ``crowding problem''. As \Perp increases, differences of distances among points will become less and less significant with respect to the length scales of the kernel in $P$ distribution, and $P$ will tend toward uniform. The forward form of KL objective function in Eq.\ \ref{kl_obj} has large cost for under-estimating probability at some point, but not for over-estimating. In other words, if $p_{ij}$ is large and $q_{ij}$ is very small, KL divergence from that term is large, but in the opposite direction of small $p_{ij}$ and large $q_{ij}$, KL is not as affected. Increasing \Perp leads to larger $\sigma_j$, and more uniform $p_{ij}$, so the easier is for the student-t distribution in low dimensional space to assign sufficient probability mass for all points. In short, increases \Perp relaxes the problem by reducing the amount of structure to be modelled so that less error is made according to $KL(P||Q)$, but one pays a cost in the second term of Eq.\ \ref{pbic}. The end result is the same: balancing between data-fit and complexity of model relative to data complexity. For this reason we will refer to $S(\Perp)$ in Eq.\ \ref{pbic} as pseudo BIC (pBIC) in the experiments. 

\subsection{Minimizing Some Description Length}

Minimum description length \citep{rissanen1978modeling} is a way to realize the Occam's razor principle for model selection. It recognizes that a model capturing any regularity in data can compress the data accordingly, hence reduced description length of the data is the description length of model plus the description length of the data compressed under the model.

Given the dimensionality gap between the original and embedding spaces, the saving in the description length is fixed, we just need to consider the extra description length paid to encode error, and try to minimize it. The $\text{KL}(P||Q)$ in Eq.\ \ref{kl_obj} is the average number of extra bits required to encode samples from $P$ using code optimized for $Q$. Since $p_{ii}$ is assumed to be $0$ in tSNE, then $M = (n^2 - n)/2$ is the number of unique pairwise probabilities. So $ M ~\text{KL}(P||Q)$ is the total number of extra bits required. On the other hand, we need encode where the extra bit length costs are paid, so we need to encode the neighborhood membership information. It takes $-log(1/n)$ to encode the identity (index) of one data point, and each data point has $\Perp$ number of neighbors on average. So there are $n (- \log(1/n))~ \Perp$ bits required to encode all neighbor identities. Taking out the factor of $M/2$, we have Eq.\ \ref{pbic}.

\section{Validation With Inferred Human Experts Preferences On Perplexity}
\label{experiments}
\vspace{-.1cm}

To validate pBIC, we infer human experts' hidden utility over perplexities by learning from their pairwise preferences over t-SNE maps on different perplexities. We show that selection by pBIC generally agrees with experts consensus. 

\subsection{Preference elicitation using Gaussian Process}
\vspace{-.1cm}
t-SNE results are precomputed from a grid of perplexities ranging from 8 to half the number of data samples, $n/2$. Users are presented with randomly selected pairs of t-SNE results from different settings. Each user chooses which map they believe to better reveal structures of the data. Users also indicate the strength of preference over a scale of four discrete choices. Once the user preferences are collected, we use a Gaussian Process (GP) model with pairwise ranking likelihood\citep{guo2010gaussian} to learn the latent utility function from collected pairwise preferences. We use the same likelihood and Laplace approximate inference as \citet{guo2010gaussian}. Using such a Bayesian framework is crucial to properly compare pBIC result against user preferences, because user preferences are uncertain given both inherent noise and potential lack of information due to insufficient sampling. Unlike in \citet{guo2010gaussian}, we do not model differences across human experts, instead pooling all their selections together. Note that for the human expert experiments, because we want to avoid introducing any complicated sequential biases, we did not use active sampling in preference elicitation, but rather random sampling on a fixed grid. To select the optimal setting using the pBIC rule in practice can be done more efficiently through Bayesian optimization or bisection search.
\subsection{Experiment results}
\begin{figure}[h]
  \centering
  \begin{subfigure}{\label{fig:digits_cl}}
      \centering
  \includegraphics[width=2.6in]{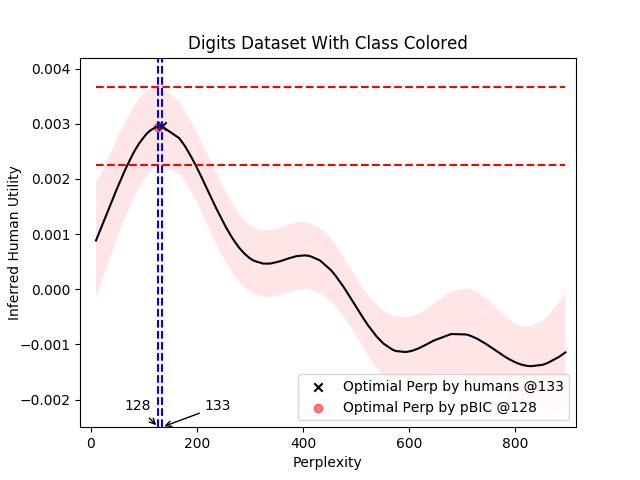} 
  \end{subfigure}  
  \begin{subfigure}{\label{fig:digits_bw}}
      \centering
  \includegraphics[width=2.6in]{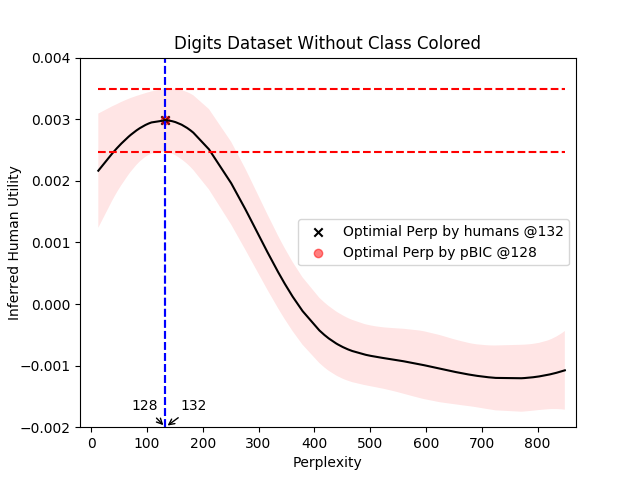}
  \end{subfigure}

  \begin{subfigure}{\label{fig:coil20_cl}}
      \centering
  \includegraphics[width=2.6in]{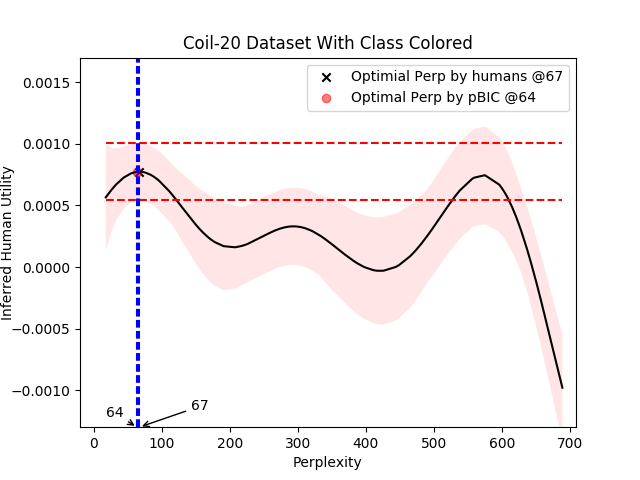}
  \end{subfigure}  
  \begin{subfigure}{\label{fig:coil20_bw}}
      \centering
  \includegraphics[width=2.6in]{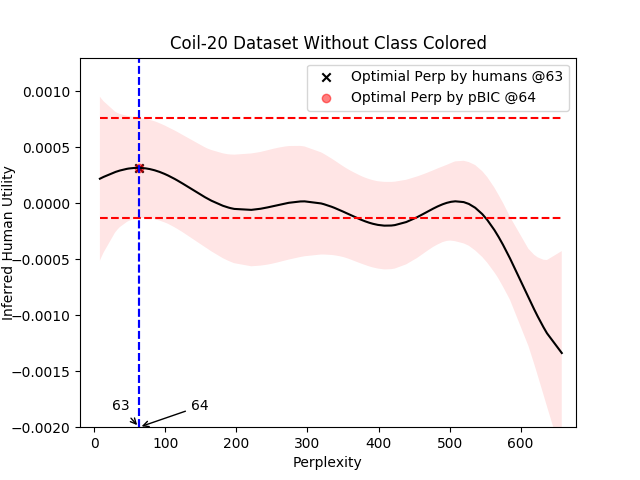}
  \end{subfigure}
  
    \begin{subfigure}{\label{fig:faces_cl}}
      \centering
  \includegraphics[width=2.6in]{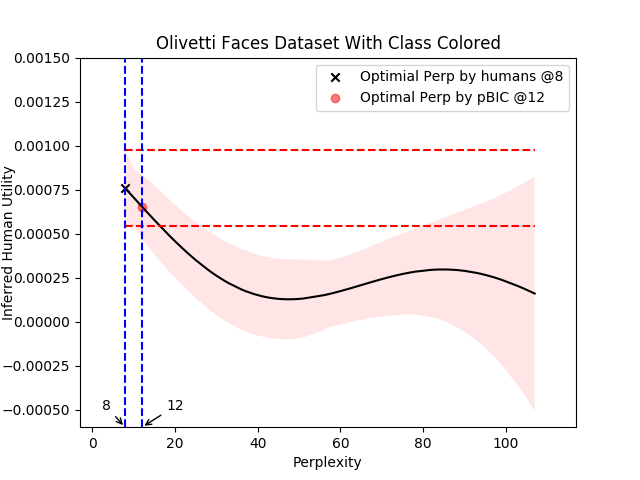}
  \end{subfigure}  
  \begin{subfigure}{\label{fig:faces_bw}}
      \centering
  \includegraphics[width=2.6in]{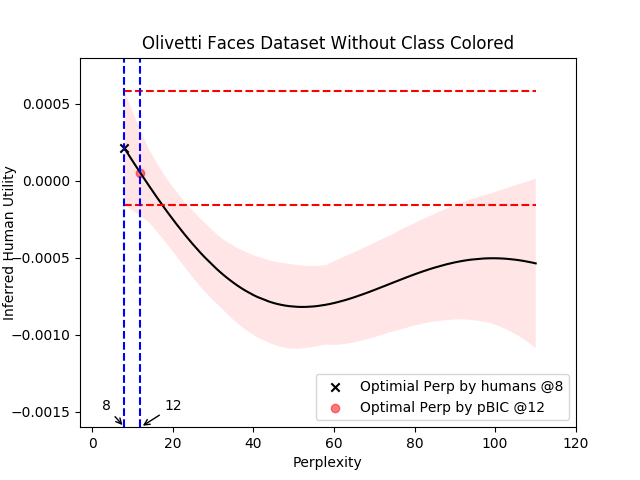}
  \end{subfigure}
  
\caption{Inferred perplexity utility functions. Three rows correspond to three datasets. 
    {\it Left Column:} experiments with class labels colored; {\it Right Column:} without class labels colored; 
    {\it Black solid lines:} GP posterior mean; {\it red shadow region:} one standard deviation in posterior variance.
    Optimal perplexity inferred from human experts marked
    as {\it a cross}, and the posterior $1\sigma$ credible region at this point is marked by {\it red dash lines}. The optimal
    perplexity from pBIC is shown as {\it a red dot}. {\it Blue dash lines} show the locations of 
    these two chosen perplexities on the $x$-axis.}
\label{main_plots}
  \end{figure}

\vspace{-.1cm}
We conducted experiments using the process described above on Handwritten Digits\footnote{\tiny{http://archive.ics.uci.edu/ml/datasets/Pen-Based+Recognition+of+Handwritten+Digits}}, Coil-20 \citep{nene1996columbia}, and Olivetti Faces dataset\footnote{\tiny{http://www.cl.cam.ac.uk/research/dtg/attarchive/facedatabase.html}}. For each dataset, preferences are collected from eight people for 30 pairs of visualizations each. Test subjects are machine learning practitioners with application or research level expertise in t-SNE. They are divided into two groups, four experts are given t-SNE maps with classes colored and the other four are presented without such information. Classes shown via colours are additional side information that can help assess the quality of embeddings, which is not available in the second group or to our pBIC method. Fig.\ \ref{main_plots} shows the results: automatic selection by pBIC and consensus implied human expert preferences are very close. When they do not match exactly, the corresponding inferred human utility at pBIC selection is so close to the peak utility that the difference is not statistically significant. In Fig.\ \ref{main_plots}, the difference is not significant if the red dot lies between the red dashed bounds, which capture $1 \sigma$ posterior credible region around the peak. See caption of Fig.\ \ref{main_plots} for more details. 

The Handwritten Digits dataset has $1797$ data points and 64 features. For either user group, pBIC picks an optimal perplexity of $128$ for this data set, whose corresponding utility is very close to the peak (Fig. \ref{fig:digits_cl} \& \ref{fig:digits_bw}).

The Coil-20 datasets contains $1440$ gray-scale pictures of $20$ objects. Pictures were taken from rotation angles and therefore the projected t-SNE maps exhibit circlar shapes (as in Fig.\ \ref{fig:pbic_y}), if the perplexity is selected appropriately. Fig.\ \ref{fig:coil20_cl} shows the optimal perplexity from pBIC is again very close to the argmax of the learned utility function. Fig.\ \ref{fig:coil20_bw}, which results from a different setting where no class label is shown to the users, has a user-preferred that is twice the pBIC-picked \Perp. However, the later is still within the $1\sigma$ confidence bounds of ther former in inferred utility, showing no significant statistical difference.

The Olivetti faces dataset has 10 profile pictures for each of the 40 people. We used random subset of 20 people (200 datapoints) to test out behaviour on a very small dataset. Again pBIC selects an optimal perplexity close to the one preferred by humans as shown in Fig.\ \ref{fig:faces_cl} and \ref{fig:faces_bw}.

\section{Conclusion}
We proposed a simple objective for automatically setting the perplexity parameter of t-SNE, making it a lot more accessible to novice users as well as reducing the risk of mis-interpreting data. We motivated the objective by relating to well known approaches in model selection and demonstrated empericially that the proposed automated method finds perplexity settings that concur with human preference on a number of problems. More formal theoretical analysis will be conducted in future research.

\bibliography{ref}



\end{document}